%% file: main.tex
\documentclass[a4paper]{article}

\usepackage{INTERSPEECH2022}

\usepackage{amsmath,graphicx}
\usepackage{multicol,multirow,array}
\DeclareMathOperator*{\argmax}{arg\,max}

\usepackage{xcolor}

\definecolor{darkspringgreen}{rgb}{0.09, 0.7, 0.27}

\title{Improving Rare Word Recognition with LM-aware MWER Training}
\name{Weiran Wang, Tongzhou Chen, Tara N. Sainath, Ehsan Variani, Rohit Prabhavalkar, \\ Ronny Huang, Bhuvana Ramabhadran, Neeraj Gaur, Sepand Mavandadi, \\Cal Peyser, Trevor Strohman, Yanzhang He, David Rybach}
\address{Google, Inc.}
\email{\{weiranwang, tongzhou, tsainath, variani\}@google.com}

\begin{document}
\maketitle

\input{0_abstract}
\input{1_intro}

\input{3_method}

\input{5_experiments}

\input{6_results}
\input{7_conclusions}

\section{Acknowledgements}
The authors would like to thank Zhong Meng for detailed discussions on the implementation differences between this work and~\cite{MengWuKandaEtAl21minimum}, and thank Erik McDermott and reviewers for the suggestions on related works.

\bibliographystyle{IEEEtran}
\bibliography{refs}

\end{document}

%% file: 0_abstract.tex
\begin{abstract}
Language models (LMs) significantly improve the recognition accuracy of end-to-end (E2E) models on words rarely seen during training, when used in either the shallow fusion or the rescoring setups.
In this work, we introduce LMs in the learning of hybrid autoregressive transducer (HAT) models in the discriminative training framework, to mitigate the training versus inference gap regarding the use of LMs. For the shallow fusion setup, we use LMs during both hypotheses generation and loss computation, and the LM-aware MWER-trained model achieves 10\% relative improvement over the model trained with standard MWER on voice search test sets containing rare words. For the rescoring setup, we learn a small neural module to generate per-token fusion weights in a data-dependent manner. This model achieves the same rescoring WER as regular MWER-trained model, but without the need for sweeping fusion weights.
\end{abstract}
\noindent\textbf{Index Terms}: hybrid autoregressive transducer, discriminative training, shallow fusion, rescoring

%% file: 1_intro.tex
\section{Introduction}
\label{sec:intro}
\vspace*{-0.5ex}

End-to-end models (E2E) are typically trained on audio-text pairs, which is a small fraction of the available text-only data. As a result, we have observed that these models do poorly on rare words~\cite{sainath2022cascaded}. In contrast, neural language models~\cite{Anjuli18,devlin2018bert,Dai2019,Keskar2019} can be trained on a much larger amount of text-only data. Performance of rare words can be improved by combining scores from the E2E and language models (LMs) during inference. These scores can be combined during beam search, known as shallow fusion~\cite{gulcehre2015using,Chorowski2017}, or rescoring a 1st-pass lattice. Given that we have an order of magnitude more text-data to train the LM, compared to the audio-text pairs used to train the E2E model, 
it is more efficient in practice to initially train the E2E model and LMs separately, and fuse them later for inference.

The training versus inference gap unavoidably arises when E2E models and LMs are trained separately.
Denote audio utterance by $X$ and label sequence by $Y$. E2E models are trained with audio-text pairs to model $P_{\text{E2E}} (Y|X)$ in the maximum likelihood estimation (MLE) training phase, while LMs are trained to model $P_{\text{LM}} (Y)$ with text-only data.
Consider the inference scenario where the scores of the two models are combined to assess different hypotheses. If the LM is already good at distinguishing certain hypotheses (with $P_{\text{LM}} (Y)$) that are hard to discriminate from the audio input, the E2E model shall not work as hard on modeling $P_{\text{E2E}} (Y|X)$ for these hypotheses, but rather spend its capacity on hypotheses that are similarly plausible according to the LM.
In addition, often we sweep the interpolation weight between the E2E and LM models on a held-out set. This step introduces additional tuning cost for complex systems (e.g., multi-lang models), and it is desirable to avoid this cost by learning the fusion weights in a data-dependent manner.

In this paper, we look to bridge the training versus inference gap,
by incorporating LMs into the E2E objective during training. This is challenging to do during MLE training, since the E2E model is trained against ground-truth labels, 
different than how LMs are incorporated during inference. In contrast, during MWER training~\cite{prabhavalkar2018,WengYuCuiEtAl20minimum,GuoTiwariDroppoEtAl20Efficient,LuMengKandaEtAl20Minimum,MengWuKandaEtAl21minimum}, we run the E2E model in beam search mode, which matches inference. Thus, it is easy to fuse scores from LMs into this beam search for training the E2E model, as well as training properly parameterized fusion weights. 
We call this technique LM-aware MWER training.

We conduct our experiments on a large-scale voice search task with a streaming Hybrid Autoregressive Transducer (HAT,~\cite{variani2020hybrid}) model, and focus on the challenge of rare word recognition.
When use LM-aware MWER in shallow fusion mode, our method significantly improve the recognition accuracy of rare words over regular MWER training, and this observation is consistent with that of~\cite{MengWuKandaEtAl21minimum} for cross-domain applications. When use in the rescoring mode, we propose to learn a small neural module for predicting the per-token fusion weights, and our method matches the WERs of using carefully tuned scalar fusion weights but avoids the need for sweeping. To the best of our knowledge this is an novel application of LM-aware MWER training in the E2E literature.

%% file: 3_method.tex
\section{LM-aware MWER training with HAT}
\label{sec:embr}
\vspace*{-.5ex}


The intuition behind discriminative training is to optimize a criterion closer to the final evaluation metric --- word error rate (WER), so that hypotheses with lower WERs have higher scores by the model, whereas hypotheses with higher WERs have lower scores. This complements the maximum likelihood estimation (MLE) training phase, in which we maximize log-probabilities of the ground truth only. While MLE has the effect of uniformly pushing down scores of all alternative label sequences than the ground truth, erroneous hypotheses may still come out at top during inference.
MWER training mitigates the training versus inference gap by adjusting scores according to WERs in the space of top hypotheses from the ASR model. 
In this section, we discuss how to perform LM-aware MWER training with an E2E model, in particular HAT, for both shallow-fusion and rescoring setups.

\subsection{Review of HAT}
\label{sec:hat}
\vspace*{-.5ex}

We use Hybrid Autoregressive Transducer (HAT), first presented in~\cite{variani2020hybrid}, as the backbone of our ASR system. HAT can be seen as a variant of RNN-T~\cite{Graves_12a} with a factorized output distribution over the vocabulary, conditioned on whether to emit a non-blank token at each frame. 
It provides a model-based estimation of an internal LM for the distribution over non-blank label sequences, whose scores can be computed by feeding zero acoustic features to the HAT decoder~\cite{variani2020hybrid,meng21}.
Using this quantity and the Bayes rule, a pseudo-likelihood sequence-level score can be derived in the density ratio method framework~\cite{morgan90, variani15, kanda17, mcdermott19}, for integration with an external language model either during a first-pass beam search or second-pass rescoring:
\begin{gather}
    \log P (Y|X) := \nonumber \\ \log P_{\text{E2E}} (Y|X) - \lambda \cdot \log P_{\text{ILM}} (Y) + \gamma \cdot \log P_{\text{ELM}} (Y) \label{eqn:final_likelihood}
\end{gather}
where $P_{\text{E2E}}$ refers to the HAT's full-sum conditional probability based on its complete output distribution (including blanks), ILM refers to the HAT model's internal LM, and ELM refers to an external LM trained only on text data. 
$(\lambda, \gamma) \ge 0$ are weights for the ILM and ELM respectively, often tuned by hand.
An intuition behind~\eqref{eqn:final_likelihood} is that the internal LM captures the bias from seeing limited text data during ASR training (with audio-text pairs), and replacing it with an external LM trained on large amount of text data effectively corrects this bias, making it 
more stable to integrate an E2E model with an external LM.

\subsection{LM-aware MWER for shallow fusion}
\label{sec:method-fusion}
\vspace*{-0.5ex}


Let $X$ be an input utterance and $Y^*$ be its ground truth label sequence.
In LM-aware MWER training for shallow fusion, we perform beam search with LMs 
to generate the top-K hypotheses. This boils down to approximately solving
\begin{align*}
    \argmax\nolimits_{Y}\; \log P (Y|X)
\end{align*}
with $\log P (Y|X)$ defined in~\eqref{eqn:final_likelihood}.
Here $\lambda \ge 0$ and $\gamma \ge 0$ are scalar weights for fusing LM scores during beam search. The above problem is solved only approximately as search is myopic and can not explore all hypotheses, and that $\log P_{\text{E2E}} (Y|X)$ is estimated using paths visited during the search.

Let $\{Y_1,\dots,Y_K\}$ be the top $K$ hypotheses produced by beam search. We make the assumption that probabilities for all possible label sequences concentrate in the top-$K$ space, and compute the re-normalized log-probabilities
\begin{align} \label{e:top-k-score}
\log \hat{P} (Y_k|X) & := \log P_{\text{E2E}} (Y_k|X) - \mu \cdot \log P_{\text{ILM}} (Y_k) \\ \nonumber
& + \nu \cdot \log P_{\text{ELM}} (Y_k)  + C, \qquad k=1,\dots,K
\end{align}
where $(\mu, \nu) \ge 0$ are fusion weights potentially different from $(\lambda, \gamma)$, and $C$ is a normalizing constant independent of $k$ such that $\sum_{k=1}^K \hat{P} (Y_k|X) = 1$. 
We use in~\eqref{e:top-k-score} the full-sum conditional probability of $Y_k$ for $P_{\text{E2E}} (Y_k|X)$. 
Incorporating both the E2E and the LM scores, $\log \hat{P} (Y_k|X)$ provides a more accurate measure of quality of $Y_k$, as opposed to using the E2E model only. Our discriminative training objective is the expected number of word errors of the top-$K$ hypotheses:
\begin{align*}
    \ell_{\text{MWER}} \left( X, Y^*, \{Y_k\}_{k=1}^K \right) := \sum\nolimits_{k=1}^K \hat{P} (Y_k|X) \cdot \text{NWE}(Y_k, Y^*)
\end{align*}
where $\text{NWE}(Y_k, Y^*)$ measures the number of word errors between $Y_k$ and $Y^*$.
As is common in the literature, for better learning stability, we minimize a composite loss
\begin{align*}
\ell_{\text{MWER}} \left( X, Y^*, \{Y_k\}_{k=1}^K \right) - \theta \cdot \log P_{\text{E2E}} (Y^*|X)
\end{align*}
over the training set, and we fix $\theta=0.04$ in this work.

The above formulation generalizes that of regular MWER, which corresponds to the special case of $\lambda=\gamma=\mu=\nu=0$, and adjusts the score of top-$K$ hypotheses according to WERs in presence of the LM to be used for inference. Thus we expect the trained E2E model to cooperate better with the LM in the shallow fusion mode.

\noindent\textbf{Related work}
Discriminative training for the acoustic model has been studied extensively for classicial ASR systems, where the LM is an integral part of the model~\cite{povey02,vesely2013sequence}.
Recent works have studied the MWER training of RNN-T and HAT. \cite{WengYuCuiEtAl20minimum} developed a technique for interpolating RNN-T's non-blank scores with with external LM scores, for hypothesis generation and alignment-based MWER loss computation.
\cite{GuoTiwariDroppoEtAl20Efficient} performed semi-on-the-fly hypothesis generation (i.e., beam search is done offline with stale model weights) with the N-best list provided by external LM rescoring, and used full-sum $\hat{P}(Y_k|X)$ in MWER loss which is adopted by subsequent works (including ours).
\cite{LuMengKandaEtAl20Minimum} performed regular MWER training of HAT (without using LMs for hypothesis generation or MWER loss), and investigated the robustness to decoding hyper-parameters in the density ratio approach, with internal and external LMs.
Our formulation in this section is the same as that of~\cite{MengWuKandaEtAl21minimum}\footnote{There exists difference in the HAT implementation in that~\cite{MengWuKandaEtAl21minimum} did not enforce a factorized distribution over blank versus non-blanks in its full softmax during training, but re-normalized the non-blank logits computed with zero acoustic features during inference to obtain the internal LM scores; our implementation consistently uses the factorized distribution (\cite{variani2020hybrid}, eqn 4) for both training and inference.}, and our results corroborate with theirs. However, we generalize the overall framework to be more versatile, and enable deep integration of LMs in the rescoring setup as discussed in the next section, which is a novel application of LM-aware MWER training in the E2E literature to 
our knowledge.

MWER training has been applied to other end-to-end models including CTC~\cite{shannon2017optimizing}, recurrent neural aligner~\cite{sak17}, and attention-based seq2seq models~\cite{prabhavalkar2018}. For the same goal of improving rare word recognition, \cite{peyser20} proposed LM-aware MWER training of a deliberation model, using external LMs trained on large text corpus similar to ours. The deliberation model has an attention-based decoder which makes predictions based on a first-pass model's hypotheses in addition to its encoder outputs. Our work differs from~\cite{peyser20} in that we perform MWER training of a first-pass model, and the choice of HAT enables us to incorporate internal LM in a principled approach with the density ratio method.






\subsection{Learnable fusion module for rescoring}
\label{sec:method-rescore}
\vspace{-0.5ex}

In latency-constrained applications, external LMs are often used in the rescoring mode, i.e., the E2E model first produces the top-$K$ hypotheses, and then external LM computes scores for each hypothesis label sequence, which are linearly combined with the E2E model scores and internal LM scores (which can be computed during the search) to determine the final scores and ranking of the top-$K$ hypotheses.
While the internal LM and external LM provide scores for each token of the hypothesis, most often the same scalar fusion weight is shared by all tokens and all utterances for each LM, and they are tuned to minimize WER on a held-out set or optimized during training \cite{variani2020neural}. 

We propose to use a learnable fusion module (LFM in short) to compute \emph{per-token} fusion weights for LM scores, in a data-dependent manner.
The LFM has a transformer architecture similar to attention-based decoder~\cite{Vaswani_17a}. Given the input hypotheses and encoder outputs, LFM performs a (causal) self-attention to the hypothesis features and a cross-attention to the acoustic features within each transformer layer, and outputs two scalars per token, used as fusion weights for internal LM score and external LM score for that token respectively. Formally, let $Y=[y_1,\dots,y_L]$ be a hypothesis of $L$ tokens for input utterance $X$, the LFM produces weights $\mu (X, Y) = [\mu_1,\dots,\mu_L]$ for internal LM, and let internal LM scores be $\log P_{\text{ILM}} (Y) = [s_1,\dots,s_L]$, then the total score contribution by internal LM is
\begin{align*}
    \mu (X, Y) \cdot \log P_{\text{ILM}} (Y) = \sum\nolimits_{l=1}^L \mu_l \, s_l.
\end{align*}
The total score contribution $\nu \cdot \log P_{\text{ELM}} (Y)$ by external LM is computed similarly. Clearly, LFM generalizes scalar fusion weights and are intuitively more flexible.
We can then plug the score contributions into~\eqref{e:top-k-score} and train the LFM in the LM-aware MWER framework. On the other hand, we set $\lambda=\gamma=0$ in beam search so that no LM is used for hypothesis generation as is the case for rescoring. Once trained, the LFW can be directly deployed to compute fusion weights without further tuning, and final scores $\log \hat{P}(Y_k|X)$, $k=1,\dots,K$ are used for re-ranking. In practice, we find that a very small LFM is sufficient to achieve similar WERs as standard LM rescoring, but without the need for careful sweeping of scalar weights.

%% file: 5_experiments.tex
\section{Experiments}
\label{sec:expts}
\vspace*{-.5ex}






\noindent\textbf{E2E Model} Our E2E model is similar to \cite{Rami21}, namely a 120M streaming HAT, which has a conformer encoder with 12 layers and attention dimension 512, and a V2 embedding decoder (i.e., the prediction network computes LM features based on two previous non-blank tokens).
The acoustic training data includes anonymized and hand-transcribed audio data covering the search, farfield, telephony and YouTube domains~\cite{narayanan2019recognizing} and has gone through multi-condition training (MTR~\cite{kim2017generation}) and random 8kHz down-sampling~\cite{li2012improving} as data augmentation. 
We also transcribe 500M unsupervised voice search utterances with a classical model and mix them with hand-transcribed data for training, with a mixing ratio of 90\% for hand-transcribed data versus 10\% for pseudo-transcribed data in each mini-batch. 

We perform beam search with a beam size of 8 for hypotheses generation in MWER training and final inference; the MWER loss is computed on the top-8 hypotheses.
MLE training takes 500K steps, while discriminative training takes additional 25K steps for the shallow fusion setup and 300K steps for LFM, with a global minibatch size of 4096 utterances.
Spec-Augment~\cite{park2020specaugment} is applied for MLE training, and turned off for hypotheses generation during MWER training following~\cite{sainath2022improving}. 

\noindent\textbf{Language Models} 
We experiment with two external LMs.
The first is a LSTM 
with 4 layers of 2048 units, and has a total of 127M parameters. The second is a conformer 
with 12 layers and attention dimension 384, and has a total of 70M parameters.
For training the LMs, each minibatch is sampled 50/50 from the transcripts of acoustic training data with a total of 150M unique transcripts, and text-only data which contains 50B utterances. The text-only data contains textual search queries from the domains of Maps, Qsession, News, Play, and Youtube, and a frequency-based pruning strategy, designed to improve rare word modeling, is implemented to adjust the probability of selecting each query~\cite{huang2022sentence}. All acoustic and LM training data is anonymized and adheres to Google AI Principles~\cite{googleaiprinciples}.


\noindent\textbf{Evaluation} 
We use both real audio and TTS generated data for evaluation. The TTS sets contain rare proper nouns (RPN) which appear less than 5 times in the training set. The test sets still abides by~\cite{googleaiprinciples}
as the original set used to select rare words is anonymized and follow the same principles. Each TTS set contains 10K utterances and covers one of five domains: Maps, News, Play, QSession, and Youtube. These TTS sets are denoted as RPN-M, RPN-N, RPN-P, RPN-Q, and RPN-Y.

We use RPN-Maps, as well as a set of 12K voice search (VS) utterances 
as development sets.
Final evaluation is performed on five sets, including RPN-N, RPN-P, RPN-Q, RPN-Y, and a side-by-side losses test set (SXS) containing 1K utterances where the quality of the E2E model transcription has more errors than a state-of-the-art conventional model~\cite{Golan16}.

For evaluation with an LMs in either shallow-fusion mode or rescoring mode, unless otherwise specified, scalar LM weights, different from the $\lambda$ and $\gamma$ used in training, are swept carefully with vizier \cite{Golovin2017} on the development sets. The weight combination that achieve best averaged WER on VS and RPN-M are used for evaluating test WERs; this criterion tries to ensures that we perform well on rare words without degrading on typical traffic. No sweeping is done in the case of LFM. 

%% file: 6_results.tex
\subsection{Results for shallow fusion}
\label{s:expts-shallow-fusion}
\vspace*{-0.5ex}

\begin{table}[t]
    \centering
    \caption{WERs (\%) of LM-aware training in shallow fusion mode. For our method, we provide in parenthesis the $(\lambda, \gamma)$ values used for training.  Unless specified, WERs are obtained with shallow fusion and scalar fusion weights are swept.}
        \label{tab:shallow-fusion}
    \vspace*{-1.5ex}
    \begin{tabular}{@{}|@{\hspace{0.01\linewidth}}c@{\hspace{0.01\linewidth}}|@{\hspace{0.01\linewidth}}c@{\hspace{0.01\linewidth}}|@{\hspace{0.01\linewidth}}c@{\hspace{0.01\linewidth}}||@{\hspace{0.01\linewidth}}c@{\hspace{0.01\linewidth}}|@{\hspace{0.01\linewidth}}c@{\hspace{0.01\linewidth}}|@{\hspace{0.01\linewidth}}c@{\hspace{0.01\linewidth}}|@{\hspace{0.01\linewidth}}c@{\hspace{0.01\linewidth}}|@{\hspace{0.01\linewidth}}c@{\hspace{0.01\linewidth}}|@{}}
            \hline
    Method & VS & RPN-M & -N & -P & -Q & -Y & SXS \\ \hline \hline
    \multicolumn{8}{|c|}{Fusion with LSTM LM (127M)} \\ \hline
       MLE training &  6.3 & 10.3 & 7.4 & 31.8 & 13.8 & 18.3 & 20.7 \\ 
       no LM in inference
       & 6.4 & 15.7 & 8.5 & 36.6 & 21.8 & 24.3 & 24.3 \\ 
                   \hline
        MWER (0.0, 0.0) & 6.0 & 10.0 & 7.4 & 31.2 & 13.2 & 18.2 & 19.7  \\
        LM-aware (0.0, 0.2) & 5.6 & 9.6 & 7.2 & 30.0 & 12.5 & 17.3 & 18.0 \\
        LM-aware (0.2, 0.3) & \bf 5.5 & 9.1 & \bf 7.1 & 28.9 & 11.9 & 16.4 & 18.1 \\
         \hspace{2.5em} no sweeping & \bf 5.5 & 9.4 & 7.2 & 29.3 & 12.5 & 16.8 & 18.3 \\
        LM-aware (0.3, 0.4) & 5.6 & 9.0 & \bf 7.1 & \bf 28.8 & 11.7 & \bf 16.0 & 17.5 \\
        \hspace{2.5em} no sweeping & 5.6 & \bf 8.9 & \bf 7.1 & 28.9 & \bf 11.6 & \bf 16.0 & \bf 17.3 \\ \hline \hline
        
            \multicolumn{8}{|c|}{Fusion with Conformer LM (70M)} \\ \hline
       MLE-training &  6.2 & 10.7 & 7.5 & 31.9 & 14.1 & 18.5 & 20.1 \\ 
       MWER (0.0, 0.0) & 5.9 & 10.3 & 7.4 & 31.2 & 13.8 & 18.5 & 19.5 \\
    LM-aware (0.2, 0.3) & 5.6 & 9.4 & 7.1 & 29.5 & 12.0 & 16.6 & 17.8 \\
    \hline
    \end{tabular}
    \vspace*{-1ex}
\end{table}

\begin{table*}[t]
    \centering
    \caption{Case study of models with conformer LM fusion on RPN sets. Red color indicates {\color{red} errors}  while blue indicates {\color{blue} corrections}.}
    \label{tab:case-study}
    \vspace*{-1.5ex}
    \begin{tabular}{@{}|@{\hspace{0.01\linewidth}}c@{\hspace{0.01\linewidth}}|c@{\hspace{0.01\linewidth}}|@{\hspace{0.01\linewidth}}c@{\hspace{0.01\linewidth}}|@{}} \hline
    Testset & Regular MWER & LM-aware MWER \\ \hline\hline
    \multirow{ 2}{*}{\vspace{3ex} \parbox{0.06\linewidth}{RPN-News}} &
    {\parbox{0.43\linewidth}{\vspace{.5ex} on offense they took a couple of steps forward when they drafted offensive tackle {\color{red} just in Pew} and offensive guard Eric Herman \\}} &  
    {\parbox{0.43\linewidth}{\vspace{.5ex} on offense they took a couple of steps forward when they drafted offensive tackle {\color{blue} Justin Pugh} and offensive guard Eric Herman \\}} \\[-1.2ex] \cline{2-3}
    
    &\parbox{0.43\linewidth}{\vspace{.5ex} {\color{red} associate a general} teams offer advice and services to individual corporate and institutional customers in Three core businesses \\} &  
    \parbox{0.43\linewidth}{\vspace{.5ex} {\color{blue} Societe generale} teams offer advice and services to individual corporate and institutional customers in Three core businesses \\} \\[-1.3ex] \hline
    
    \multirow{ 2}{*}{\vspace{1.5ex} {\parbox{0.06\linewidth}{RPN-Maps}}} &
    {\parbox{0.43\linewidth}{\vspace{.5ex} Hardeeville recycling art evil South Carolina \\}} &	
    {\parbox{0.43\linewidth}{\vspace{.5ex} Hardeeville recycling {\color{blue} Hardeeville} South Carolina \\}} \\[-1.2ex] \cline{2-3}
    
    & {\parbox{0.43\linewidth}{\vspace{.5ex} where is {\color{red} long you're viewing \\}}} &
    {\parbox{0.43\linewidth}{\vspace{.5ex} where is {\color{blue} longyearbyen \\}}}  \\[-1.2ex] \hline
    
    \multirow{ 2}{*}{\vspace{1.5ex} {\parbox{0.06\linewidth}{RPN-Youtube}}} 
    & {\parbox{0.43\linewidth}{\vspace{.5ex} {\color{red} West Side gun} Pat LaFontaine\\}}	& 
    {\parbox{0.43\linewidth}{\vspace{.5ex} {\color{blue} Westside Gunn} Pat LaFontaine\\}}  \\[-1.1ex] \cline{2-3}
    
    & \raisebox{-1ex}{\parbox{0.43\linewidth}{{\color{red} our body yassi's} always alright\\[-1.1ex]}}	& 
    \raisebox{-1ex}{\parbox{0.43\linewidth}{{\color{blue} arbani yasiz} Always Alright\\[-1.1ex]}} \\ \hline
    \end{tabular}
    \vspace*{-1ex}
\end{table*}

We first demonstrate our method in the shallow fusion scenario. During training, we set $\lambda=\mu$ and $\gamma=\nu$ for simplicity, i.e., the same LM weights are used for hypotheses generation and loss calculation; this reduces the number of parameters to be tuned. 


We explore different $(\lambda, \gamma)$ combinations and report the shallow fusion WERs of models trained with LSTM LM in Table~\ref{tab:shallow-fusion} (top section). The MLE-train model, used to initialize LM-aware training, is a baseline and this model achieves 6.3\% and 10.3\% on VS and RPN-M respectively (as a reference, the WERs without shallow fusion for inference are 6.4\% and 15.7\%, the gap demonstrates the effectiveness of LMs). The optimal scalar weights from sweeping are around 0.2 for internal LM and 0.3 for external LM, and these values provide guidance for setting $\lambda$ and $\gamma$ for LM-aware training.
With regular MWER, which is a special case of our method with $\lambda=0$ and $\gamma=0$, the development set WERs reduce to 6.0\% and 10.0\%, showing the benefit of a discriminative training criterion. As another special case of our method, we set $\lambda=0.0$ and $\gamma=0.2$ so that only the external LM is used for training ($0.2$ is the optimal scalar weight for external LM fusion for the MLE-trained model when no internal LM is used), and this model significantly reduced development set WERs to 5.6\% and 9.6\%, demonstrating the effectiveness external LM-aware training. Finally we train a few models with non-zero internal LM weights, around $(\lambda=0.2, \gamma=0.3)$. These model behave similarly and we report results for two combinations $(0.2, 0.3)$ and $(0.3, 0.4)$. While these models provides small improvements for VS, they further reduce RPN-M WER to 9.0\%, as compared to the 9.6\% achieved by $(0.0, 0.2)$. With internal LM scores subtracted to correct the bias from E2E training data, we are able to use larger external LM weight to obtain further WER gain. 

We have also tried to evaluate the models without re-sweeping the LM weights, i.e., we use the same fusion weights $(\lambda,\gamma)$ used in training for evaluation, and WERs for the last two models are given in Table~\ref{tab:shallow-fusion}. We observe small WER degradation from the those obtained with re-sweeping, consistent with the fact that the weights found by re-sweeping do not deviate too much from the (0.2, 0.3) combination we started with. This indicates that if we start with good estimates of the 
weights, we may 
perform only light sweeping around them after 
training.


Comparing different methods on the five test sets, we observe that LM-aware MWER training improves over the baseline and regular MWER consistently and uniformly. With LSTM LM fusion, MLE-trained model has an averaged WER of 18.40\% while regular MWER yields an averaged WER of 17.94\%. LM-aware training with $(\lambda=0.3, \gamma=0.4)$ significantly reduces the averaged WER to 16.22\%, enjoying a 9.6\% relative WER reduction from regular MWER.

\begin{table}[t]
    \centering
    \caption{Rescoring WERs (\%) with the 70M conformer LM. LFM uses freezed E2E model from regular MWER training.
    }
    \label{tab:rescoring}
    \vspace*{-1.5ex}
    \begin{tabular}{@{}|@{\hspace{0.01\linewidth}}c@{\hspace{0.01\linewidth}}|@{\hspace{0.01\linewidth}}c@{\hspace{0.01\linewidth}}|@{\hspace{0.01\linewidth}}c@{\hspace{0.01\linewidth}}||@{\hspace{0.01\linewidth}}c@{\hspace{0.01\linewidth}}|@{\hspace{0.01\linewidth}}c@{\hspace{0.01\linewidth}}|@{\hspace{0.01\linewidth}}c@{\hspace{0.01\linewidth}}|@{\hspace{0.01\linewidth}}c@{\hspace{0.01\linewidth}}|@{\hspace{0.01\linewidth}}c@{\hspace{0.01\linewidth}}|@{}}
            \hline
    Method & VS & RPN-M & -N & -P & -Q & -Y & SXS \\ \hline
        regular MWER & 5.9 & 10.7 & 7.7 & 31.6 & 14.4 & 19.4 & 19.9 \\ 
        no LM in inference
        & 6.0 & 14.3	& 8.3 & 35.2 & 20.4 & 23.3 & 22.9 \\
        \hline 
        \hline
        LFM\ \  4.2M  & 5.8 & 11.0 & 7.8 & 32.7 & 15.1 & 19.7 & 18.4 \\ 
        LFM\ \  6.2M  & 5.6 & 11.1 & 7.8 & 33.3 & 15.4 & 19.6 & 18.3 \\  \hline\hline
        \parbox{0.34\linewidth}{LM-aware for E2E \\[-.3ex]
        $\lambda$=$\gamma$=0, $\mu$=0.2, $\nu$=0.3}
        & 5.6 & 10.4 & 7.6 & 31.0 & 13.9 & 18.8 & 19.1  \\ \hline
    \end{tabular}
    \vspace*{-1ex}
\end{table}

We repeat some of the experiments with the conformer LM and report results in the bottom section of Table~\ref{tab:shallow-fusion}. In general conformer LM fusion results are slightly worse due to the smaller LM size, but the relative merits of methods are quite consistent. LM-aware MWER training leads to 8.7\% relative reduction of averaged WER from regular MWER on the set sets.
In Table~\ref{tab:case-study} we provide case study of decoding results of models trained by regular MWER and LM-aware MWER on RPN sets.

\subsection{Results for rescoring}
\label{s:expts-rescoring}
\vspace*{-0.5ex}

In the rescoring setup, we freeze the E2E model obtained from regular MWER training, and only update the parameters of a small LFM for 300K steps. Our LFMs consist of three transformer layers, and we experiment with two attention dimensions 256 and 320, resulting in a total of 4.2M and 6.2M parameters respectively. 
We use the conformer LM in this section.

We present the rescoring WERs of LFM modules in Table~\ref{tab:rescoring}. As a baseline, the regular MWER-trained model achieves 5.9\% and 10.7\% on VS and RPN-M respectively (as a reference, the WERs without rescoring for inference are 6.0\% and 14.3\%). 
Compared to the baselines, LFM models perform clearly better on VS, with the 6.2M LFM achieving 5.6\% on VS. When sweeping the scalar weights for non-LFM models, we observe different operational points that trade off WERs on VS and RPN-M, but did not find a model performing as well on VS; this demonstrates the flexibility of learnable fusion weights. On the other hand, LFM models perform worse on RPNM than swepted scalar weights. The test set results are consistent with those on dev sets, in the sense that LFM models outperform baselines on the real audio set SXS, but have higher WERs on the TTS-generated RPN sets. 
We hypothesize this is because LFM is data dependent, and given our training data is more representative of typical voice search traffic, LFM models are naturally biased towards data from matching domain (and it is even more so for the larger LFM), and perhaps adding TTS data into training will improve results on RPN sets.

\begin{figure}[t] 
    \centering
    \begin{tabular}{@{}c@{\hspace{0.06\linewidth}}c@{}}
    \includegraphics[width=0.43\linewidth, page=1]{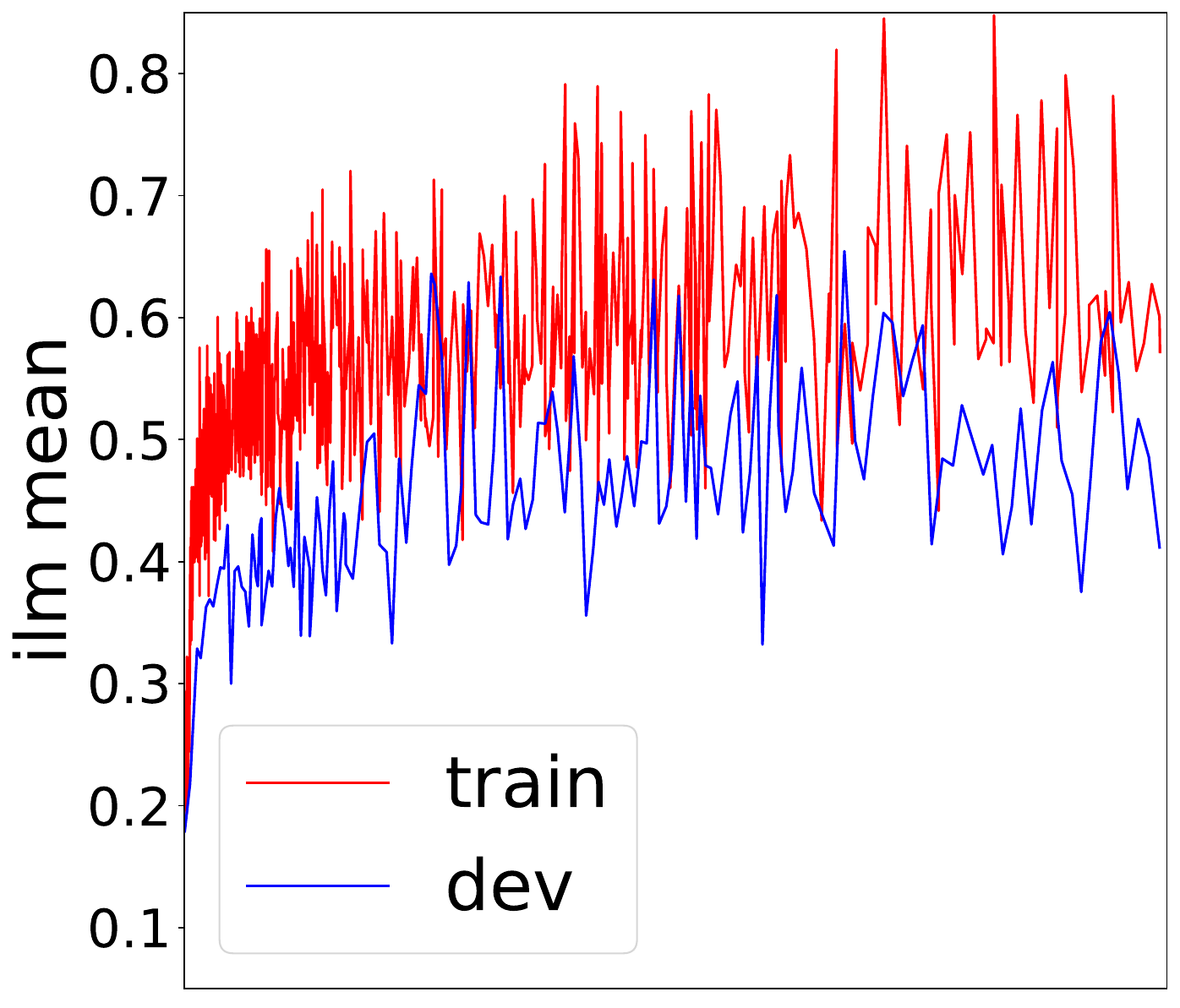} &
    \includegraphics[width=0.43\linewidth, page=1]{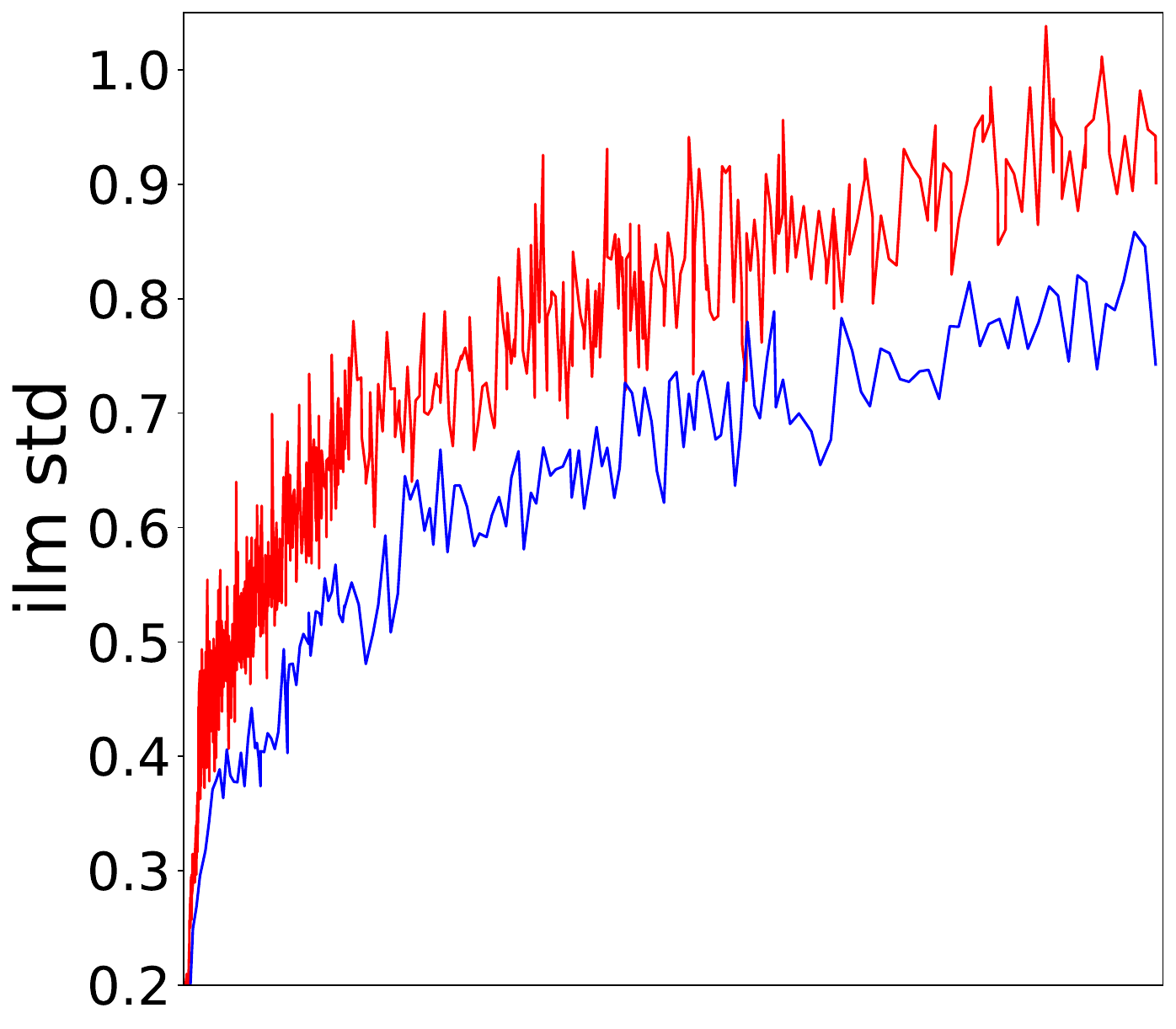} \\
    \includegraphics[width=0.43\linewidth, page=1]{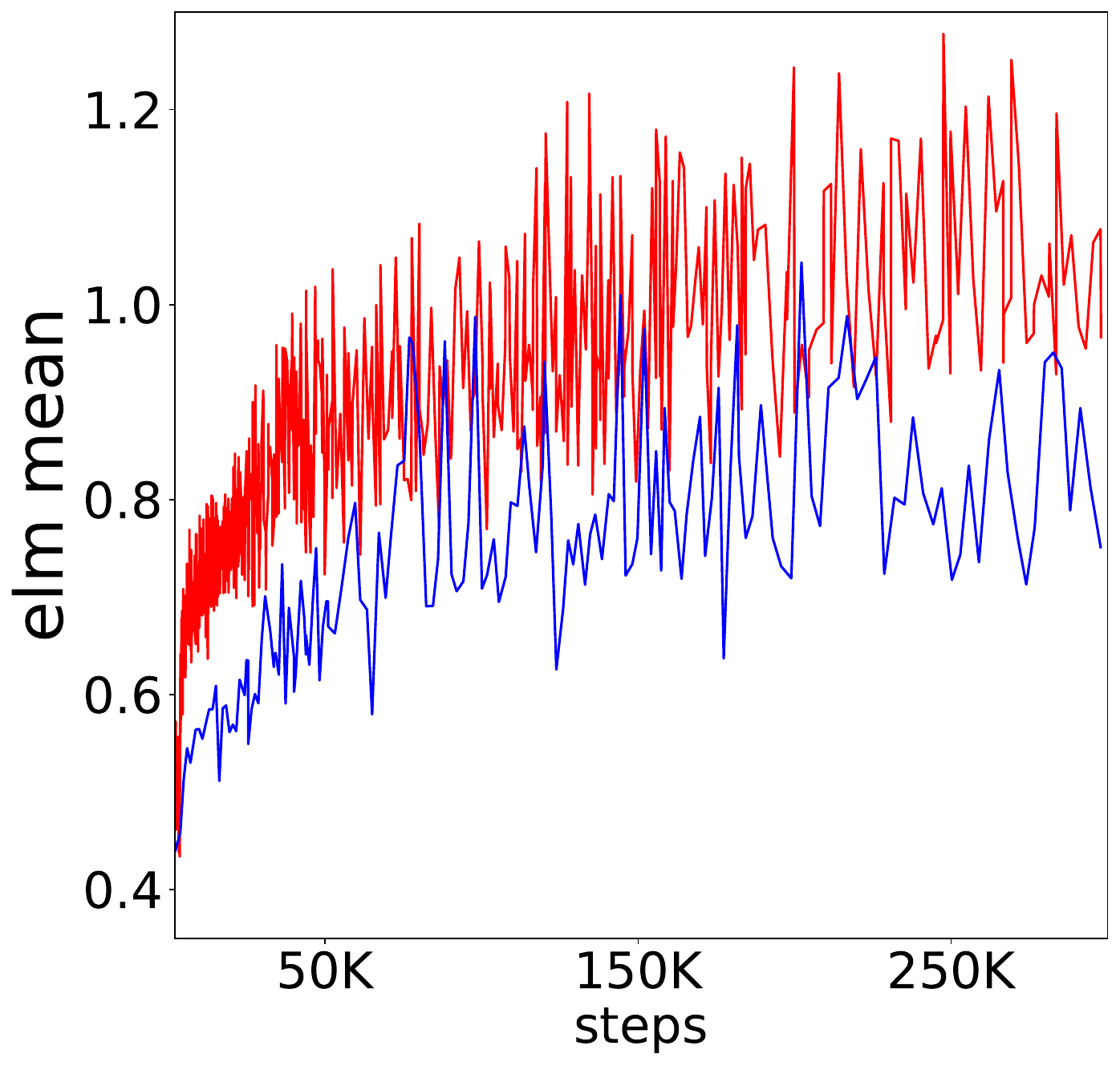} &
    \includegraphics[width=0.43\linewidth, page=1]{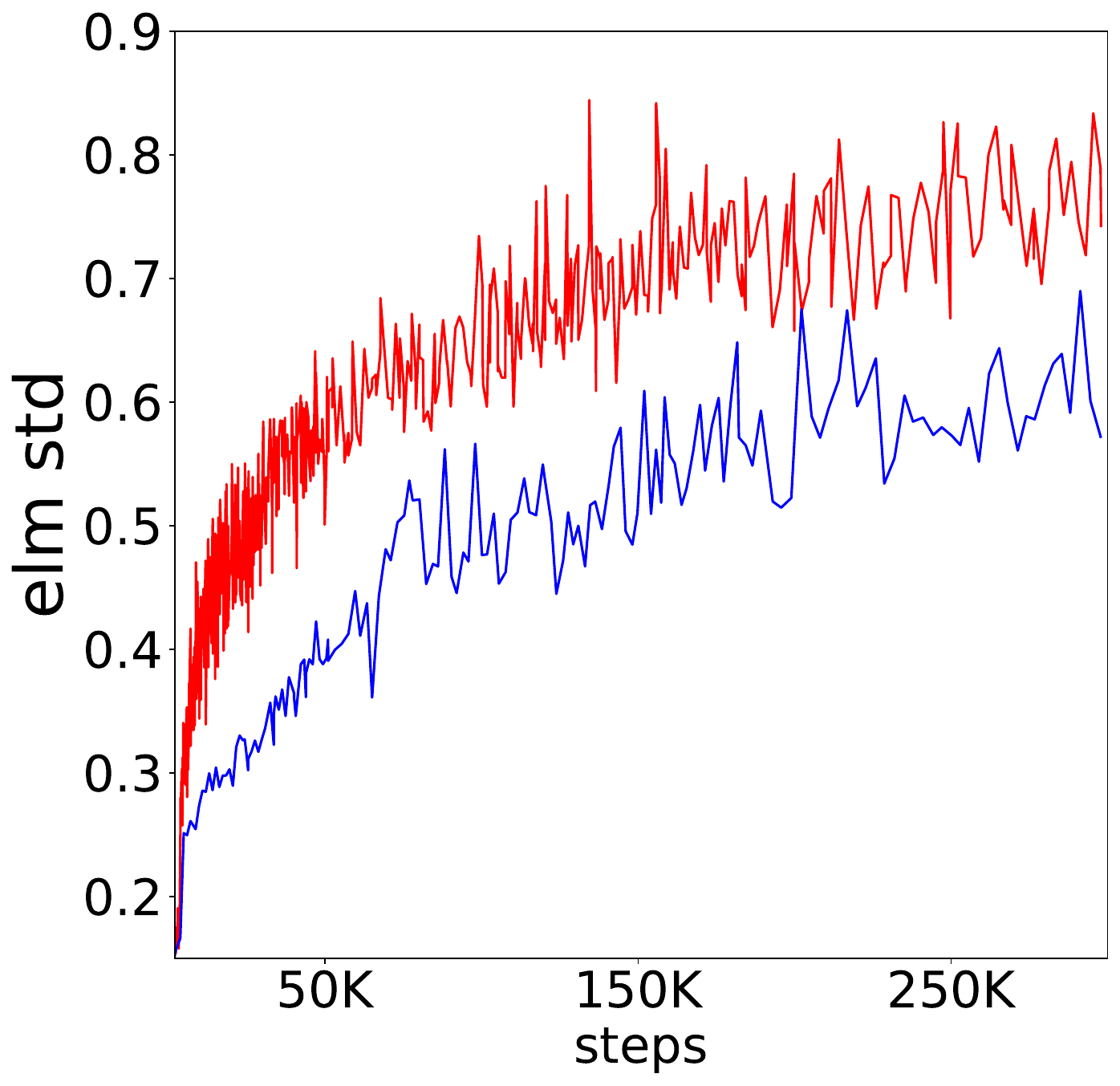}
    \end{tabular} 
    \vspace*{-2.5ex}
    \caption{Mean and standard deviation of internal and external LM fusion weights on training and dev sets during training.}
    \label{fig:fusion_weights}
    \vspace*{-2ex}
\end{figure}

Overall, LFMs achieve an averaged test WERs of 18.74\% (4.2M model) and 18.88\% (6.2M model), similar to the 18.60\% by the swept regular MWER-trained model (these three models use the same E2E model checkpoint).
We provide the learned internal and external LM fusion weights by the 6.2M LFM module during training in Figure~\ref{fig:fusion_weights}. Observe that the learned weights have on average larger magnitude than the swept scalar weights (around 0.3 and 0.4 for internal and external LMs respectively), with significant variance across utterances and tokens.

We have also trained a LM-aware MWER model from the MLE initialization, using the method of Sec~\ref{sec:method-fusion} with $\lambda=0, \gamma=0, \mu=0.2$ and $\nu=0.3$, tailored for the rescoring setup.
With 120M E2E model parameters updated during training (much more than the LFM size), this model achieves 5.6\% and 10.4\% on the dev sets as shown in Table~\ref{tab:rescoring} (last row).
The nice gain from updating the E2E model for rescoring
motivated us to jointly train the E2E model and LFM, but then we observed the E2E model to degrade quickly with low quality hypotheses when the LFM is randomly initialized. It is a future direction to develop a joint training mechanism that constrains the E2E model to not deviate too much from initialization.

%% file: 7_conclusions.tex
\section{Conclusions}
\label{sec:conclusions}
\vspace*{-.5ex}

We have developed a flexible framework that performs training of the E2E model or a small learnable fusion module, in the presence of LMs to be used during inference, and thus more effectively integrates text information into ASR models. Our method leads to improved recognition accuracy on rare words in the shallow fusion mode, and avoids expensive sweeping in the rescoring mode.
In the future, we would like to extend the learnable fusion module into the shallow fusion mode, to produce data-dependent fusion weights for hypothesis generation. Given the large gains from discriminative training on the top hypotheses, another promising direction is to generalize our framework to use dense RNN-T lattices~\cite{prabhavalkar2021less}.